\documentclass{article} % For LaTeX2e
\usepackage{colm2024_conference}

\usepackage{microtype}
\usepackage{hyperref}
\usepackage{url}
\usepackage{booktabs}
\usepackage{graphicx}
\usepackage{multirow}
\usepackage{amssymb}
\usepackage{bbding}
\usepackage{arydshln}
\usepackage{booktabs}
\usepackage{amsmath}
\usepackage{wrapfig}
\usepackage{bm}
\newcommand{\prompt}[1]{{\footnotesize \textsf{#1}}}

\title{Look Before You Leap: Problem Elaboration Prompting Improves Mathematical Reasoning in Large Language Models}
\author{Haoran Liao, Jidong Tian, Shaohua Hu, Hao He, Yaohui Jin \\
MoE Key Lab of Artificial Intelligence, AI Institute, Shanghai Jiao Tong University \\
\texttt{\{liaohaoran,frank92,hushaohua,hehao,jinyh\}@sjtu.edu.cn}
}

\colmfinalcopy % Uncomment for camera-ready version, but NOT for submission.
\begin{document}

\maketitle

\begin{abstract}
\vspace{-5pt}
Large language models (LLMs) still grapple with complex tasks like mathematical reasoning. 
Despite significant efforts invested in improving prefix prompts or reasoning process, the crucial role of problem context might have been neglected.
Accurate recognition of inputs is fundamental for solving mathematical tasks, as ill-formed problems could potentially mislead LLM's reasoning. 
In this study, we propose a new approach named Problem Elaboration Prompting~(PEP) to enhance the mathematical capacities of LLMs.
Specifically, PEP decomposes and elucidates the problem context before reasoning, therefore enhancing the context modeling and parsing efficiency. 
Experiments across datasets and models demonstrate promising performances: (1) 
PEP demonstrates an overall enhancement in various mathematical tasks. For instance, with the GPT-3.5 model, PEP exhibits improvements of 9.93\% and 8.80\% on GSM8k through greedy decoding and self-consistency, respectively.
(2) PEP can be easily implemented and integrated with other prompting methods. (3) PEP shows particular strength in handling distraction problems~(example in Fig.~\ref{fig::distraction-example}).
\vspace{-10pt}
\end{abstract}

\begin{figure}[h]
    \centering
    \includegraphics[width=.95\textwidth]{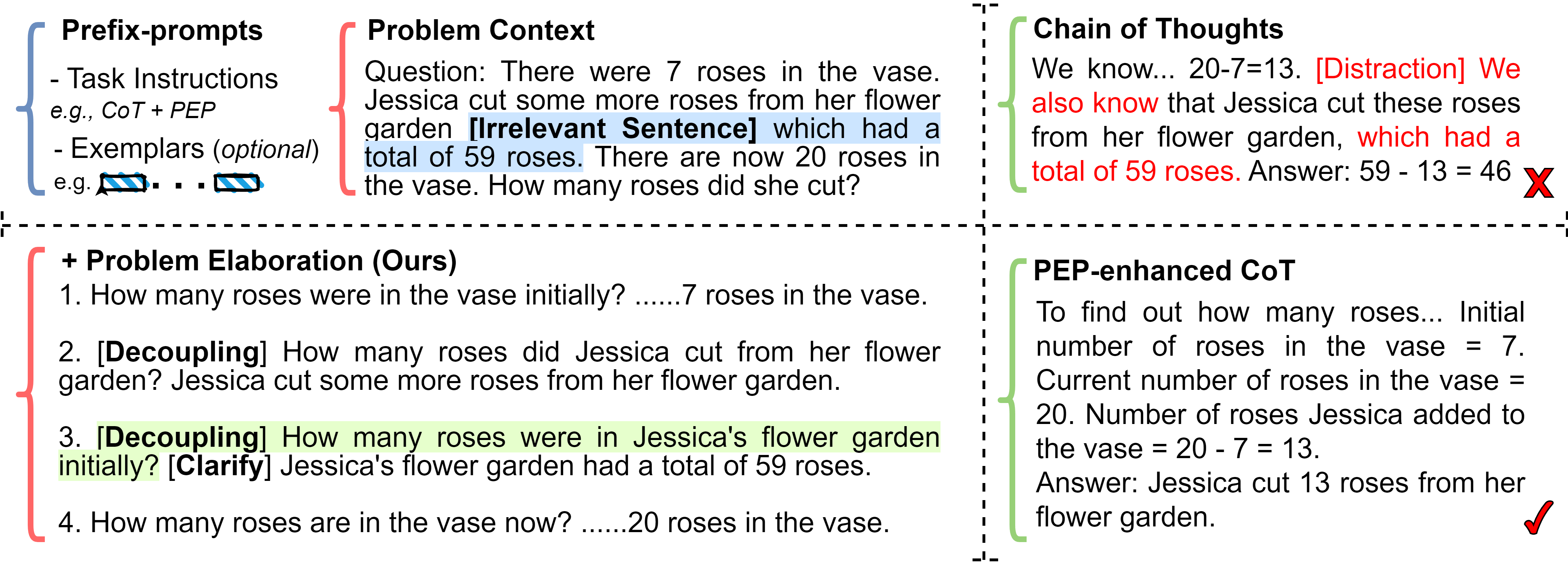}
    \caption{
    We proposed Problem Elaboration Prompting~(PEP) for enhancing problem context, thereby improving subsequent reasoning. As depicted in the example, PEP decouples spurious relationships and refines statements, preventing downstream distraction errors.}
    \label{fig::distraction-example}
\end{figure}

\section{Introduction}
% large-scale pretrained language models
Recent large language models~(LLMs), such as the GPT-3 model family with 175 billion parameters~\citep[\textit{inter alia}]{brown2020language}, have demonstrated remarkable performance across various NLP tasks. 
Chain-of-thought (CoT) prompting~\citep{wei2022chain,kojima2022large} successfully elicits reasoning behavior and emergent abilities~\citep{Wei2022EmergentAO} by explicitly guiding the model to generate intermediate rationales step by step, further promoting the development of artificial general intelligence~(AGI). 
Despite the success, performing multi-hop and compositional reasoning for complex tasks like mathematical solving can still face challenges~\citep{Hendrycks2021MeasuringMP,Lewkowycz2022SolvingQR}, even when the required knowledge is limited to the scope of primary school~\citep{Cobbe2021TrainingVT}.

\begin{figure*}[t!]
  \centering
  \includegraphics[width=\textwidth]{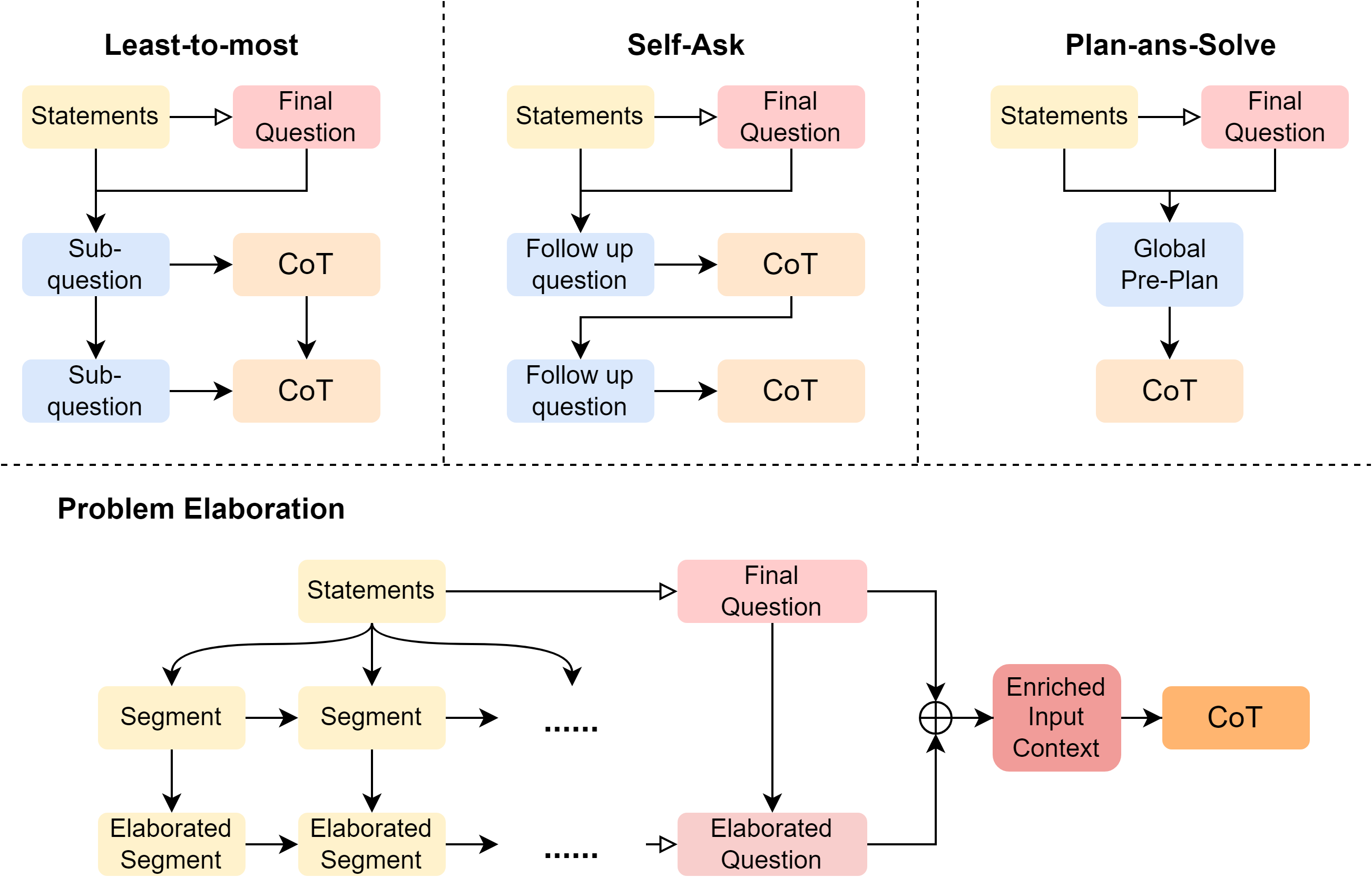}
  \caption{
  An overview of the proposed PEP and other problem-related methods. Rather than creating sub-questions or plans to guide subsequent reasoning, PEP focuses on clarifying and enriching the problem context, i.e., PEP can be integrated with these methods.}
  \label{fig:overview}
\end{figure*}

One area of research aims to improve the quality of reasoning outputs through diverse rationale decoding strategies~\citep{Wang2022SelfConsistencyIC,Suzgun2022FollowTW} and iteration-based answer refinement~\citep{Saunders2022SelfcritiquingMF,Kim2023LanguageMC,Zheng2023ProgressiveHintPI}.  
Considering the sensitivity of the model to inputs~\citep{lu2022fantastically,Wang2022TowardsUC,Shi2023LargeLM}, another area of research focuses on augmenting the robustness of prefix-prompts~\citep{Fu2022ComplexityBasedPF,Wang2022RationaleAugmentedEI,Shao2023SyntheticPG}.
However, the role of problem has been overlooked. 

Most studies assume that the provided information is concise and relevant, while in real-world situations, the inputs could be ill-formed.
For instance, LMs can be easily distracted by irrelevant context~\citep{Shi2023LargeLM} and often struggle to capture intricate implied implicatures~\citep{Ruis2022LargeLM,Chan2023ChatGPTEO}.
Besides, 
even when the problem is well-formed, it may still be complex and unsuitable for the LM's comprehension. 
For instance, although the language model can be knowledgeable, it may encounter difficulties in identifying what knowledge are required to answer the question~\citep{Bian2023ChatGPTIA} or correctly integrating intermediate rationales to generate the overall solution~\citep{Press2022MeasuringAN}.

Several works have noticed the crucial role of problem, suggesting pre-processing before reasoning. Least-to-Most~(L2M)~\citep{Zhou2022LeasttoMostPE} proposes to decompose the final asked question into simpler sub-questions, Plan-and-Solve (PaS) ~\citep{Wang2023PlanandSolvePI} requires a preliminary global plan, while Self-ask~\citep{Press2022MeasuringAN} suggests a dynamic asking strategy before each step of generating rationales. However, these methods primarily focus on decomposing questions or creating guidance, without understanding or discernment of the problem itself. Therefore, they could also potentially be misled by ill-formed problems.

In this study, we introduce a new method named \textbf{P}roblem \textbf{E}laboration \textbf{P}rompting~(PEP), which involves decomposing and elucidating the problem context prior to reasoning. Our method aims to clarify the problem description and enhance the context modeling, rather than creating specific guidance.  
We illustrated an overview of PEP in Fig~\ref{fig:overview}, along with a comparison with other problem-related prompting methods. 
Specifically, PEP adopts a human-like thought process that emphasizes the importance of thoroughly comprehending the problem's conditions and requirements before engaging in reasoning: \textit{look before you leap}.
PEP is also inspired by previous researches on semantic parsing, which suggest parsing the given problem into specific representations, such as Python code or condensed symbols~\citep{Gao2022PALPL,Chen2022ProgramOT,Hu2023ChainofSymbolPE}, to facilitate subsequent reasoning.

We conduct evaluations of the proposed approach on four mathematical datasets and additionally investigate its performance in addressing the distraction problems~\citep{Shi2023LargeLM}.
Both zero-shot and few-shot PEPs demonstrate an overall enhancement across datasets, models and answer types, employing greedy decoding and self-consistency settings. For instance, with GPT-3.5, we observed an improvement of 9.93\% and 8.80\% on GSM8k~\citep{Cobbe2021TrainingVT} with greedy decoding and self-consistency, separately. When using ChatGPT, PEP achieves SOTAs with 98.23\% on SingleEq~\citep{KoncelKedziorski2015ParsingAW}, and 88.7\% on SVAMP~\citep{Patel2021AreNM}.

In summary, our contributions are listed below:
\begin{enumerate}
    \item 
    We propose a new method, Problem Elaboration Prompting (PEP), for enhancing LLM's reasoning. It is easy to implement and integrate other prompting methods.    
    \item We evaluate PEP through extensive experiments using both open-source LMs and GPT model family, exploring the role of problem context for reasoning.
    \item We demonstrate PEP is effective in mitigating the distraction problem, indicating the promising prospect in dealing with ill-formed problems of other types.
\end{enumerate}

\section{Related Works}

\subsection{Emergent Abilities and Prompting}
With the large-scale unsupervised pre-training technique, LLMs~\citep[\textit{inter alia}]{brown2020language,Chowdhery2022PaLMSL,touvron2023llama} can perform new tasks conditioning on few in-context inputs via mimicking the format of demonstrations~\citep{webson2022prompt,Min2022RethinkingTR,Pan2023WhatIL}. Instruction tuning further improves the generalization of prompts via training a language model to follow the general instructions rather than specific examples~\citep{Wei2021FinetunedLM,Ouyang2022TrainingLM,Chung2022ScalingIL}. 

Instead of directly generating the final answer, chain-of-thought prompting~(CoT) methods~\citep{wei2022chain,Creswell2022SelectionInferenceEL,Jung2022MaieuticPL} suggest guiding LLMs to reach the final answer through a step-by-step process, eliciting the emergent reasoning ability~\citep{Wei2022EmergentAO,Chan2022DataDP,Prystawski2023WhyTS}. \citet{Zelikman2022STaRBR} also proposed to use prompts to augment the trainset with rationales.
However, such methods typically necessitate an LLM larger than 100B~\citep{wei2022chain}, making it difficult to directly apply on small models~\citep{zhang2023multimodal}.
\subsection{Improving CoT Reasoning}
% \subsection{Improving CoT Reasoning.}
Various techniques have been proposed to improve the standard CoT~\citep{Wang2022SelfConsistencyIC,Suzgun2022FollowTW,Saunders2022SelfcritiquingMF,Gao2022PALPL,Kim2023LanguageMC,Yao2023TreeOT}, most with a particular focus on controlling the decoding process and validating intermediate answers. 
Significant efforts have also been made to improve the prompts selection and ensemble~\citep{Fu2022ComplexityBasedPF,Lu2022DynamicPL,Zhou2022LargeLM,Shao2023SyntheticPG}, to enhance the generation robustness. Besides, increasing the diversity of prompts, questions, or rationales shows notable advantages for the reasoning~\citep{Zhang2022AutomaticCO,Wang2022RationaleAugmentedEI}. 

PEP shares similarity with the study conducted by 
\citet{Lampinen2022CanLM}, which explores the impact of explanations in context.
However, it focus on the post-answer explanation, while we focus on elaborating the problem.
% \subsection{Collaborating CoT with agents.}
% Recent works attempt to introduce the concept of agents to enhance the reasoning, explicitly prompting the LLM with external tools~\citep{Gao2022PALPL} to play different roles in undertaking decomposed tasks ~\citep{Shinn2023ReflexionLA,GenerativeAgents,Voyager}. It typically includes planning, cooperation, competition, and interaction with the environment~\citep{Xi2023TheRA}. 
% % Leveraging external tools~\citep{Gao2022PALPL} and exploring the decision-making process~\citep{Jung2022MaieuticPL,Yao2023TreeOT} have also shown promising improvements.
% Exploring the decision-making process~\citep{Jung2022MaieuticPL} with specific agents has also shown promising improvements, including tree-structure~\citep{Yao2023TreeOT} and graph-structure~\citep{multichoice,graph1}.
PEP can also be regarded as a type of LLM-based agent system, sharing the concepts of analysis and decomposition~\citep{Shinn2023ReflexionLA,GenerativeAgents,Voyager}. While they primarily focus on observing environment and reflecting the feedback~\citep{Zheng2023ProgressiveHintPI, creator,mathchat,selfcheck}, PEP focuses on comprehending and parsing the context.

\section{Problem Elaboration Prompting}

A beneficial cognitive paradigm of humanity is the ability of thoughtful contemplation, as exemplified by a common figurative: \textit{look before you leap}. In this work, we propose to simulate such a thinking process by prompting the model to elaborate on the problem before generating or reasoning. It is crucial for the LLM to ensure it has a thorough understanding of the problem space, encompassing the given conditions and the asked question, avoiding a hasty generation and exploration within the answer space~(Table~\ref{tab:hasty_demo}).

\begin{table}[t!]
    \centering
    \small
    \begin{tabular}{p{.9\linewidth}}
    \toprule
    \textbf{Q}: Brian's friend Bobby has 5 fewer than 3 times as many video games as Brian does. If Brian has 20 video games \textcolor{cyan}{but lost 5 right before the comparison was made}, how many does Bobby have? \\
    \midrule
    \textbf{Standard CoT:} \\
    \textbf{A}: \textcolor{red}{Three times as many video games as Brian has is 3*20=60 video games}. Five fewer than 3 times as many video games as Brian has is 60-5=55 video games. \\\\
    Final answer: 55. 
    \textbf{(Wrong: ignore the precondition placed at end.)}\\
    \midrule
    \textbf{PEP + CoT} \\
    (+\textbf{Elaboration}): \\
    Segment 1: Bobby has 5 fewer than 3 times as many video games as Brian does. Rephrased: Bobby's number of video games is 3 times Brian's number minus 5. Equation: B = 3B - 5\\
    Segment 2: Brian has 20 video games. Equation: B = 20\\
    Segment 3: Brian lost 5 video games right before the comparison was made. Equation: B = 20 - 5\\
    \\
    \textbf{A}: Using segment 2's equation, we know that Brian has 20 video games. \textcolor{blue}{Using segment 3's equation, we know that Brian lost 5 video games, leaving him with 15 video games}. We can substitute this value into segment 1's equation and solve for Bobby's number of video games. B = 3(15) - 5, B = 40. \\\\Final answer: 40. 
    (\textbf{Correct: process conditions in suitable order.})\\
    \bottomrule
    \end{tabular}
    \caption{Recognizing the problem is imperative
for reasoning. As illustrated, CoT wrongly engages in hasty reasoning~(\textcolor{red}{red}), neglecting the prerequisite conditions placed afterwards~(\textcolor{cyan}{cyan}). Meanwhile, PEP utilizes conditions in the correct order~(\textcolor{blue}{blue}).
    }
    \label{tab:hasty_demo}
\end{table}

\subsection{Method Formulation}
Given the problem context denoted by $P=(X,y_0)\in D$, where $X = \{x_1, x_2, \cdots,x_i\}$ represent the statements and $y_0$ represents the asked question, prompting methods attempt to leverage the strength of LLMs to solve the task by generating rationale $ y_i = f_M(X,y_0;p_t,\hat{P}_k|y_{<i})$ step by step until reaching the final answer $y_t$. Specifically, $p_t$ denotes zero-shot instructions and $\hat{P}_k$ indicate $k$ concatenated exemplars $(\hat{X}_k,\hat{y}_{k,0})$. Note that such prompting methods do not modify the LLM $M$.

In the PEP-aid language model, we suggest to pre-process the given problem $P$ by decomposing and elucidating the context into smaller and more concise segments to enhance LLMs' comprehension: $P'=(\{x_1', x_2', \cdots,x_m'\},y_0')= f_M(X,y_0;p_e)$, where $m\geq i$ and $p_e$ is a specific instruction. Then, the LLM can continue its reasoning by $y_i = f_M(X,y_0;p_t,\hat{P}_k|P',y_{<i})$ utill reaching $y_t$. Thus, PEP can be easily combined with previous prompting methods.

\subsection{Designing Principles} 
The design principle of problem elaboration in this study consists of two aspects: (i) \textbf{decomposing}: breaking down the original sentences into distinct segments to disentangle complex or intertwined concepts; (ii) \textbf{elucidating}: providing explanations or rephrasing the segments in a manner that is more conducive to the model's understanding.%, thereby facilitating downstream reasoning. 

The concept of \textit{decomposition} is widely spread in many previous works.
Except for introduced problem-related methods, recent community adopts a decomposition approach in different fields to apply LMs to solve complex problems~\citep{wei2022chain, Khot2022DecomposedPA, Liang2023EncouragingDT, Hong2023MetaGPTMP}, bridging the compositionality gap of powerful LMs~\citep{Press2022MeasuringAN}.
In contrast, PEP takes a different approach by breaking down the entire problem into simpler segments, rather than creating sub-questions or decomposing the reasoning process. It focuses on organizing and clarifying the existing information from the problem.

Meanwhile, it could be beneficial to elucidate the segments following decomposition, as it elicits the model to organize existing information in a comprehensive view of the problem and recognize the underlying implicatures of the question. Moreover, it introduces diversity into the context, which has been demonstrated to enhance reasoning~\citep{Zhang2022AutomaticCO,Wang2022RationaleAugmentedEI}, thereby mitigating the risk of relying on specific words or descriptions as shortcuts. A break-down analysis of our designed principles is presented in Sec.~\ref{sec::ablation}.

\subsection{Prompts Generation} 
Since the elaboration can be diverse, We evaluate and select the instruction of zero-shot PEP based on a subset of GSM8k-train of 200 points~(see Sec~\ref{sec::prompt-selection}). 
The selected instruction is ``\textit{Decompose the given question into smaller segments, elucidating each segment as you rephrase it.}''
We further adopt the exemplars from \cite{Zhou2022LeasttoMostPE} and adjusted them for different methods for fairness. All instructions and exemplars can be found in the appendix~\ref{sec::prompt}.

To investigate the behavior of PEP, we randomly selected 200 instances from the experiments with a manual categorization. PEP primarily utilizes question-answer pairs, declarative sentences, and interrogative sentences to review and examine the problem context.
There are around 10.5\% of instances that can be mixed with sub-questions, planning instructions, or intertwined rationales. We find it primarily occurs when the questions themselves contain instructions or options.

\section{Experiment}
\subsection{Setup}

\begin{wraptable}{r}{0.3 \textwidth}
\vspace{-10pt}
\centering
\resizebox{0.3 \textwidth}{!}{
\begin{tabular}{ccc}
\toprule
Dataset & Num & Answer \\ \midrule
SingleEq & 508 & free \\
GSM8k & 1319 &   free \\
AQuA & 254 &   options \\
SVAMP & 1000  & free \\
GSMIC-1k & 1000 &   free \\
 \bottomrule
\end{tabular}
}
\caption{Datasets Stats.}
\label{table::dataset_info}
\vspace{-10pt}
\end{wraptable}

\paragraph{Datasets.} We evaluate PEP on four elementary datasets, with a focus on mathematical reasoning: 
(1) \textbf{SingleEq}~\citep{KoncelKedziorski2015ParsingAW}, (2) \textbf{GSM8k}~\citep{Cobbe2021TrainingVT}, (3) \textbf{SVAMP}~\citep{Patel2021AreNM}, (4) \textbf{AQuA}~\citep{Ling2017ProgramIB}.
Furthermore, we investigate the distraction problem using GSMIC~\citep{Shi2023LargeLM}: 
we randomly sampled 500 examples separately for 2-step problems and m-step problems, denoted by (5) \textbf{GSMIC-1k}. The details are shown in Table~\ref{table::dataset_info}.

\paragraph{Baselines \& Prompts.} 
We evaluate PEP with two elementary answer types: (1) Chain-of-Thoughts~(\textbf{CoT})~\citep{kojima2022large} for textual reasoning, and
(2) Program-of-Thoughts~(\textbf{PoT})~\citep{Chen2022ProgramOT} for code-based reasoning. 
Three problem-related methods are compared to PEP: (3) Least-to-Most~(\textbf{L2M})~\citep{Zhou2022LeasttoMostPE} , (4) Plan-and-Solve (\textbf{PaS}) ~\citep{Wang2023PlanandSolvePI} and (5) \textbf{Self-ask}~\citep{Press2022MeasuringAN}.
To investigate the distraction problem, we also adopt (6) \textsc{\textbf{Irr-Inst.}} suggested by \citet{Shi2023LargeLM}.
All instructions and exemplars can be found in the appendix~\ref{sec::prompt}.

\paragraph{Language Models \& Decoding} We conduct our main experiments on four open-source LMs (1) two \texttt{LLama2} models~\citep{Touvron2023Llama2O} and (2) two \texttt{Mistral} models~\citep{Jiang2023Mistral7,Jiang2024MixtralOE}. We also employ two GPT models (3) ``\texttt{text-davinci-003}''~(\texttt{davinci}) and (4) recent released ``\texttt{gpt-3.5-turbo-0125}''~(\texttt{turbo})~\citep{brown2020language,ouyang2022training,OpenAI2023GPT4TR}. 
We evaluate the performance with both greedy decoding and self-consistency decoding~(SC)~\citep{Wang2022SelfConsistencyIC} to ensure the reproducibility of experiments. 

\subsection{Main Results}

% Comparison between different models
\begin{table*}[t!]
\centering
\resizebox{.9\textwidth}{!}{
\begin{tabular}{cccccccc}
\toprule
 &  &  & \multicolumn{4}{c}{Dataset} &  \\ \cmidrule(lr{0pt}){4-7}
\multirow{-2}{*}{Model} & \multirow{-2}{*}{AnsType} & \multirow{-2}{*}{\begin{tabular}[c]{@{}c@{}}PEP\\(ours)\end{tabular}} & SingleEq & GSM8k & AQuA & SVAMP & \multirow{-2}{*}{Average} \\ \midrule

& \multirow{2}{*}{\textsc{CoT}}  & \XSolidBrush &  73.82 &  29.72 & 21.26 & 48.1 & \multirow{2}{*}{\color[HTML]{2EA121}(+0.33)}\\
\multirow{-2}{*}{\begin{tabular}[c]{@{}c@{}}LLama2\\ \texttt{7B-hf-chat}\end{tabular}} &  & \Checkmark & 71.87 & 27.32& 23.62 &  51.4&  \\ \midrule
& \multirow{2}{*}{\textsc{CoT}}  & \XSolidBrush &  80.91 & 41.47 & 24.41 & 59.7& \multirow{2}{*}{\color[HTML]{2EA121}(+0.68)}\\
\multirow{-2}{*}{\begin{tabular}[c]{@{}c@{}}LLama2\\ \texttt{13B-hf-chat}\end{tabular}}  & & \Checkmark & 78.54 & 42.0 & 27.17 & 61.5& \\ \midrule

\multirow{2}{*}{\begin{tabular}[c]{@{}c@{}}Mistral-7B\\ \texttt{Instruct-v0.2}\end{tabular}} & \multirow{2}{*}{\textsc{CoT}} & \XSolidBrush  & 79.33 & 44.49 & 27.35 & 66.5 & \multirow{2}{*}{\color[HTML]{D83931}(-1.04)} \\
 & & \Checkmark & 78.74 & 41.71 & 27.95 & 65.1 &  \\ \midrule

\multirow{2}{*}{\begin{tabular}[c]{@{}c@{}}Mistral-8x7B\\ \texttt{Instruct-v0.1}\end{tabular}} & \multirow{2}{*}{\textsc{CoT}} & \XSolidBrush  & 90.16 & 68.61 & 45.67 & 77.9 & \multirow{2}{*}{\color[HTML]{2EA121}(+0.21)}\\
& & \Checkmark &  89.96 & 70.74 & 44.31 & 78.2 &  \\ \midrule

\multirow{4}{*}{
\begin{tabular}[c]{@{}c@{}}GPT-3.5\\ \texttt{text-davinci} \\
\texttt{-003}\end{tabular}} 
& \multirow{2}{*}{\textsc{CoT}}  & \XSolidBrush & 87.40 & 50.19 & 38.19 & 72.1  & \multirow{2}{*}{\color[HTML]{2EA121} (+4.71)}\\ 
 & & \Checkmark & 90.35 & 60.12 & 40.55 & 75.7  & \\ 
 \cmidrule(lr{0pt}){2-8}
 & \multirow{2}{*}{\textsc{PoT}} & \XSolidBrush & 95.67 & 63.84 & 29.13 & 78.3  & \multirow{2}{*}{\color[HTML]{2EA121} (+1.27)}\\%66.74 \\
 &  & \Checkmark & 96.46 & 64.75 & 29.53 & 81.3  & \\
\midrule 
 \multirow{4}{*}{\begin{tabular}[c]{@{}c@{}}ChatGPT\\ \texttt{gpt-3.5-}\\\texttt{turbo-0125}\end{tabular}} & \multirow{2}{*}{\textsc{CoT}} & \XSolidBrush & 96.46 & 82.56 & 58.66 & 78.6  & \multirow{2}{*}{\color[HTML]{2EA121} (+0.57)} \\ 
 &  & \Checkmark & 97.05 & 82.03 & 56.3 & 83.2  & \\
 \cmidrule(lr{0pt}){2-8}
 & \multirow{2}{*}{\textsc{PoT}} & \XSolidBrush & 96.46 & 77.48 & 47.64 & 81.4 & \multirow{2}{*}{\color[HTML]{2EA121} (+1.33)}\\ 
 &  &  \Checkmark& 96.06 & 79.68 & 47.64 & 84.9  &  \\ 
\bottomrule
\end{tabular}
}
\caption{Accuracies~(x100) of proposed PEP against zero-shot CoT and PoT. 
% All results are with greedy decoding~(GD).
}
\label{tab::main-greedy}
\end{table*}

\begin{table*}[t!]
\centering
\resizebox{.9\textwidth}{!}{
\begin{tabular}{cccccccc}
\toprule
 &  & & \multicolumn{4}{c}{Dataset} &  \\ \cmidrule(lr{0pt}){4-7}
\multirow{-2}{*}{Model} & \multirow{-2}{*}{AnsType} & \multirow{-2}{*}{\begin{tabular}[c]{@{}c@{}}PEP\\(ours)\end{tabular}} & SingleEq & GSM8k & AQuA & SVAMP & \multirow{-2}{*}{Average} \\ \midrule
 & \multirow{3}{*}{\textsc{CoT}+SC} & \XSolidBrush & 87.99 & 52.31 & 39.76 & 71.7  & 62.94 \\
 & & \Checkmark & 89.57 & 61.11 & 39.76 & 73.6  & 66.01 \\
\multirow{-3}{*}{\begin{tabular}[c]{@{}c@{}}GPT-3.5\\ \texttt{text-davinci} \\\texttt{-003}\end{tabular}} & & & {\color[HTML]{2EA121} (+1.58)} & {\color[HTML]{2EA121} (+8.80)} & (+0.00) & {\color[HTML]{2EA121} (1.90)} & {\color[HTML]{2EA121} (+3.07)} \\ \midrule
\multirow{3}{*}{\begin{tabular}[c]{@{}c@{}}ChatGPT\\ \texttt{gpt-3.5-}\\\texttt{turbo-0125}\end{tabular}} & \multirow{3}{*}{\textsc{CoT}+SC} & \XSolidBrush & 97.83 & 87.57 & 67.72 & 86.3  & 84.85 \\
 & & \Checkmark & 98.23 & 88.48 & 67.32 & 88.7 &  85.68 \\
 &  & & {\color[HTML]{2EA121} (+0.40)} & {\color[HTML]{2EA121} (+0.91)} & {\color[HTML]{D83931} (-0.40)} & {\color[HTML]{2EA121} (+2.40)}  & {\color[HTML]{2EA121} (+0.83)} \\ \bottomrule
\end{tabular}
}
\caption{Solve rates~(x100) of proposed PEP against self-consistency CoT. The temperature is 0.7 and each answer are voted from 20 samples.}
\label{tab::main-sc}
\end{table*}

The main results are presented in Table~\ref{tab::main-greedy} and Table~\ref{tab::main-sc}, showing proposed PEP's performances on various LMs, answer types and decoding strategies. 
We only test PoT on two GPT models, as its templates might not be designed for these smaller LMs.
Considering the token cost, we validate the self-consistency CoT on two GPT models. The comparison and integration with other problem-related methods are shown in table~\ref{tab::main-meths}, using greedy decoding and \texttt{turbo-0125}, while verifying two different few-shots settings.

\paragraph{PEP performs well in a variety of situations.} Overall, PEP outperforms the standard CoT in most cases, with the exception of \texttt{Mistral-7B}, but demonstrating improvements for \texttt{Mistral-8x7B}. 
The performance in PoT and self-consistency settings further validates PEP's effectiveness and adaptability. The enhancement in PoT could be attributed to PEP simplifying parsing difficulties, thereby facilitating code generation~\citep{Jiang2023SelfplanningCG}.
It is also noteworthy that PEP performs remarkably well on \texttt{davinci}, achieving improvements of 9.93\% and 8.80\% in greedy search and self-consistency, respectively. 

\paragraph{PEP can be effectively integrated with other prompting methods.} Unlike the mentioned problem-related methods, PEP aims to enhance the original problem, making it compatible for integration. As shown in Table~\ref{tab::main-meths}, incorporating PEP improves the performance of these problem-related methods in both k=1 and k=4 few-shot settings, while the combination with the standard CoT also ranks highly in performance. Besides, we observed Self-ask underperforms when k=1. It's likely because one example might fail to elicit the dynamic QA process and LMs could abruptly terminate after generating follow-up questions.

\begin{table*}[t!]
\centering
\resizebox{.9\textwidth}{!}{
\begin{tabular}{ccccccccc}
\toprule
& & & \multicolumn{4}{c}{Dataset} &  \\ \cmidrule(lr{0pt}){4-7}
\multirow{-2}{*}{Methods} & \multirow{-2}{*}{K-shots} & \multirow{-2}{*}{\begin{tabular}{c}
     PEP  \\
     (ours)
\end{tabular}}& SingleEq & GSM8k & AQuA & SVAMP & \multirow{-2}{*}{Average} \\ \midrule
\multirow{2}{*}{CoT}   & \multirow{2}{*}{k=1} & \XSolidBrush & \underline{97.64} & \underline{80.59} & 54.33 & 81.3 & \multirow{2}{*}{\color[HTML]{2EA121}(+0.48)}\\
& & \Checkmark & 97.24 & 79.38 & 54.72 & \underline{84.4} \\ 
\cmidrule(lr{0pt}){3-8}

\multirow{2}{*}{L2M} & \multirow{2}{*}{k=1} & \XSolidBrush & 96.85 & $\mathbf{82.11}$ & \underline{57.09} & 83.4 & \multirow{2}{*}{\color[HTML]{2EA121}(+0.65)} \\
&  & \Checkmark & $\mathbf{98.03}$ & 80.44 & \underline{57.09} & $\mathbf{84.5}$ & \\\cmidrule(lr{0pt}){3-8}
\multirow{2}{*}{PaS*} & \multirow{2}{*}{k=1} & \XSolidBrush & 93.9 & 77.63 & 54.33 & 78.7 & \multirow{2}{*}{\color[HTML]{D83931}(-0.01)} \\
&  & \Checkmark & 94.09 & 77.63 & 55.51 & 77.3 &  \\ \cmidrule(lr{0pt}){3-8}
\multirow{2}{*}{Self-Ask*} &  \multirow{2}{*}{k=1}& \XSolidBrush & 89.57 & 61.11 & 40.94 & 74.2 & \multirow{2}{*}{\color[HTML]{2EA121}(+7.63)}\\
& & \Checkmark & 95.08 & 59.29 & $\mathbf{58.27}$ & 83.7 & \\
\midrule
\multirow{2}{*}{CoT}  & \multirow{2}{*}{k=4} & \XSolidBrush & $\mathbf{98.03}$ & \underline{82.94} & 56.69 & 82.3 & \multirow{2}{*}{\color[HTML]{2EA121}(+0.99)}\\
&  & \Checkmark & $\mathbf{98.03}$ & $\mathbf{83.93}$ & \underline{59.06} & 82.9 \\ \cmidrule(lr{0pt}){3-8}
\multirow{2}{*}{L2M} & \multirow{2}{*}{k=4} &\XSolidBrush &  \underline{97.64} & 81.65 & 57.87 & 82.5 & \multirow{2}{*}{\color[HTML]{2EA121}(+1.57)} \\
&  &  \Checkmark & 97.44 & 81.8 & $\mathbf{61.02}$ & \underline{85.7} & \\\cmidrule(lr{0pt}){3-8}
\multirow{2}{*}{PaS*} & \multirow{2}{*}{k=4} & \XSolidBrush &  96.26 & 79.76 & 57.48 & 77.9 &  \multirow{2}{*}{\color[HTML]{2EA121}(+1.50)} \\
&  &\Checkmark &  96.85 & 81.58 & 57.48 & 81.5 &  \\ \cmidrule(lr{0pt}){3-8}
\multirow{2}{*}{Self-Ask*} & \multirow{2}{*}{k=4} & \XSolidBrush & 94.49 & 78.7 & 53.54 & 80.0 & \multirow{2}{*}{\color[HTML]{2EA121}(+3.20)}\\
& & \Checkmark & 95.47 & 81.43 & 55.91 & $\mathbf{86.7}$ & \\
\bottomrule
\end{tabular}
}
\caption{Comparison and integration of PEP with problem-related methods. We utilized exemplars from \cite{Zhou2022LeasttoMostPE} and adjusted them for other methods*~(details in Sec~\ref{sec::prompt}). The best and second-best results are highlighted in bold and underlined, respectively.}
\label{tab::main-meths}
\end{table*}

\subsection{Distraction Problem}
\label{sec::distraction}

% LLMs may encounter difficulties when handling ill-formed problems. 
One particular challenge of ill-formed problems is known as \textit{distraction problem}~\citep{Shi2023LargeLM}: irrelevant sentences can distract LMs to generate errors. These sentences can be completely irrelevant or relevant to the problem but should have no impact on inference. 
We tested PEP using a subset of GSMIC~\citep{Shi2023LargeLM}. Two metrics are utilized: (1) \textit{micro accuracy}: averaged accuracy per example, and (2) \textit{macro accuracy}: averaged accuracy per base problem. \textit{Norm} is the accuracy normalized by scores on base problems, measuring how a method is affected by the distractors.

\paragraph{PEP effectively mitigates the distraction problems.} As shown in Table~\ref{tab::gsmic-main}, PEP surpasses CoT and L2M of both zero-shot and one-shot settings, in micro- and macro- metrics, indicating superior performance in addressing such ill-formed issues. Beyond overall accuracy, PEP also exhibits enhanced robustness as evidenced by norm accuracy. From the macro perspective, the improvements and stability of PEP are also remarkable.

\paragraph{PEP performs well when prompted with prior knowledge.} When the model is consciously prompted to ignore irrelevant content for the given problem, referred to as~\textsc{Irr-Inst.}~\citep{Shi2023LargeLM}, we observed significant improvements in CoT, L2M and PEP. Despite this, PEP still outperforms the 0-CoT and 1-L2M by achieving larger improvements for most cases. PEP particularly excels in 2-step problems and norm accuracies. However, its performance on macro accuracies is inferior, potentially due to a conflict between \texttt{Irr-Inst.} and the one-shot exemplar.

\begin{table*}[t!]
\centering
\resizebox{\textwidth}{!}{
\begin{tabular}{cccccccccc}
\toprule
\multirow{2}{*}{Method}
 & 
\multirow{2}{*}{\begin{tabular}{c}
PEP  \\
(ours) 
 \end{tabular}} & \multicolumn{4}{c}{\textbf{Micro Accuracy}} & \multicolumn{4}{c}{\textbf{Macro Accuracy}} \\ \cmidrule(lr{0pt}){3-6}\cmidrule(lr{0pt}){7-10}
& & 2 st & $\geq$2 st & Overall & Norm & 2 st & $\geq$2 st & Overall & Norm \\ \midrule
\multirow{3}{*}{0-CoT} & \XSolidBrush & 75.4 & 78.0 & 76.7 & 89.08 & 31.67 & 40.0 & 35.0 & 48.61 \\
 & \Checkmark & 84.2 & 81.8 & 83.0    & 92.32 & 56.67 & 30.0 & 46.0    & 61.33 \\
 && {\color[HTML]{2EA121}(+8.8)} & {\color[HTML]{2EA121}(+3.8)} & {\color[HTML]{2EA121}(+6.3)} & {\color[HTML]{2EA121}(+3.2)} & {\color[HTML]{2EA121}(+25.0)} & {\color[HTML]{D83931} (-10.0)} & {\color[HTML]{2EA121}(+11.0)} & {\color[HTML]{2EA121}(+12.7)} \\
 
\multirow{3}{*}{1-CoT} & \XSolidBrush & 84.2 & 81.2 & 82.7 & 90.98 & 46.67 & 32.5 & 41.0 & 47.67 \\
 & \Checkmark & 87.8 & 81.8 & 84.8 & 95.07 & 56.67 & 37.5 & 49.0 & 61.25 \\ 
 && {\color[HTML]{2EA121}(+3.6)} & {\color[HTML]{2EA121}(+0.6)} & {\color[HTML]{2EA121}(+2.1)} & {\color[HTML]{2EA121}(+4.1)} & {\color[HTML]{2EA121}(+10.0)} & {\color[HTML]{2EA121}(+5.0)} & {\color[HTML]{2EA121}(+8.0)} & {\color[HTML]{2EA121}(+13.6)} \\
 
 \multirow{3}{*}{1-L2M} & \XSolidBrush & 84.6 & 80.8 & 82.7 & 90.98 & 51.67 & 27.5 & 42.0 & 51.22 \\
 & \Checkmark & 83.2 & 84.0 & 83.6 & 93.62 & 50.0 & 32.5 & 43.0 & 58.11 \\ 
 &&{\color[HTML]{D83931} (-1.4)} & {\color[HTML]{2EA121}(+3.2)} & {\color[HTML]{2EA121}(+0.9)} & {\color[HTML]{2EA121}(+2.6)} & {\color[HTML]{D83931} (-1.7)} & {\color[HTML]{2EA121}(+5.0)} & {\color[HTML]{2EA121}(+1.0)} & {\color[HTML]{2EA121}(+6.9)}\\
 
 \midrule
\multicolumn{9}{l}{+ \textsc{Irr-Inst.}~\citep{Shi2023LargeLM}} \\
\multirow{2}{*}{0-CoT} & \XSolidBrush & 80.6 & 85.8 & 83.2 & 97.65 & 58.33 & 55.0 & 57.0 & 72.15 \\
 & \Checkmark & 88.2 & 87.2 & 87.7 & 98.76 & 61.67 & 50.0 & 57.0 & 76.0 \\
 &&  {\color[HTML]{2EA121}(+7.6)} & {\color[HTML]{2EA121}(+1.4)} & {\color[HTML]{2EA121}(+4.5)} & {\color[HTML]{2EA121}(+1.1)} & {\color[HTML]{2EA121}(+3.3)} & {\color[HTML]{D83931} (-5.0)} & (0.0) & {\color[HTML]{2EA121}(+3.8)}  \\
 
 \multirow{2}{*}{1-CoT} & \XSolidBrush &  87.4 & 89.4 & 88.4 & 94.55 & 68.33 & 62.5 & 66.0 & 83.54\\
 & \Checkmark & 90.6 & 83.6 & 87.1 & 96.89 & 71.67 & 57.5 & 66.0 & 78.57\\
 && {\color[HTML]{2EA121}(+3.2)} & {\color[HTML]{D83931} (-5.8)} & {\color[HTML]{D83931} (-1.3)} & {\color[HTML]{2EA121}(+2.3)} & {\color[HTML]{2EA121}(+3.3)} & {\color[HTML]{D83931} (-5.0)} & (0.0) & {\color[HTML]{D83931} (-5.0)}\\

  \multirow{2}{*}{1-L2M} & \XSolidBrush &   85.8 & 85.8 & 85.8 & 96.3 & 61.67 & 50.0 & 57.0 & 74.03 \\
 & \Checkmark & 88.0 & 85.0 & 86.5 & 96.86 & 61.67 & 45.0 & 55.0 & 69.62\\
 &&  {\color[HTML]{2EA121}(+2.2)} & {\color[HTML]{D83931} (-0.8)} & {\color[HTML]{2EA121}(+0.7)} & {\color[HTML]{2EA121}(+0.6)} & (0.0) & {\color[HTML]{D83931} (-5.0)} & {\color[HTML]{D83931} (-2.0)} & {\color[HTML]{D83931} (-4.4)}\\
 
 \bottomrule

\end{tabular}
}
\caption{Micro and macro accuracies (×100) on GSMIC-1k. We follow the metrics and instructions suggested in~\citep{Shi2023LargeLM}. 
}
\label{tab::gsmic-main}
\end{table*}

\begin{figure}[ht]
    \centering
\includegraphics[width=\textwidth]{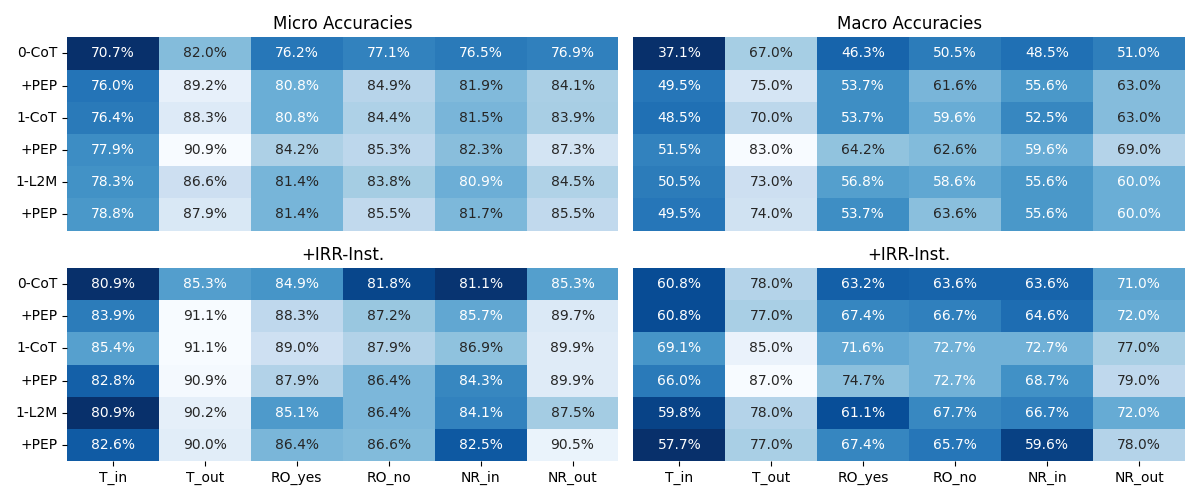}
    \caption{Breakdown accuracies w.r.t. irrelevant sentence factors~(T: Topic, RO: Role Overlap, NR: Num. Range). Lower accuracy suggests the model is more sensitive to that factor.
}
    \label{fig:gsmic-factors}
\end{figure}

\subsection{Ablation Study}
\label{sec::ablation} 

\paragraph{Breakdown analysis of distracting factors.} We evaluated the performance of CoT, L2M and PEP under various distracting factors, as depicted in Fig~\ref{fig:gsmic-factors}. PEP significantly outperforms the basic baselines on almost all factors, both in micro and macro accuracies, indicating its potential benefits for downstream reasoning by aiding in problem context recognition and parsing. When prompted by the \texttt{Irr-inst.}, PEP shows consistent improvements for most cases, except for 1-CoT. Overall, PEP demonstrates better improvements in handling out-of-distribution distraction factors, specifically (1) off-topic sentences, (2) non-overlapping role names, and (3) out-of-range numbers. The impact of these factors is more pronounced on the macro- than the micro- metric.

\begin{wraptable}{r}{0.5\textwidth}
\centering
\resizebox{.5\textwidth}{!}{
\begin{tabular}{ccccc} \toprule
Method & GSM8k & SVAMP  & GSMIC & Overall \\ \midrule
\textsc{CoT} & 82.56 & 78.6 & 76.7 & 79.60 \\ 
\textsc{PEP} & 82.03 & 83.2  & 83.0 & 82.67 \\ \midrule
\textsc{Dec} & 82.79 & 83.6  & 79.0 & 82.50 \\
\textsc{Elu} & 81.80 & 83.1  & 82.2 & 82.57 \\
\textsc{ETD} & 80.14 & 82.5  & 81.1 & 81.59 \\ \bottomrule
\end{tabular}
}
\caption{Ablation of PEP components. }
\label{tab::ablation-pep}
\vspace{-8pt}
\end{wraptable}
% \end{table}

\paragraph{Break-down analysis of PEP components.} 
We verify two components in PEP: (1) \textsc{Dec}: decomposing only, and (2) \textsc{Elu}: elucidating only, and (3) \textsc{EtD}: elucidating first then decomposing. As shown in Table~\ref{tab::ablation-pep},
both \textsc{Dec} and \textsc{Elu} are required for PEP, with performance varying across datasets. On GSM8k and SVAMP, \textsc{Dec} even outperforms PEP, while \textsc{Elu} is more effective on GSMIC. Besides, \textsc{EtD} consistently performed worse, suggesting that the coordination and operating order of components are also crucial for PEP.

\subsection{Error Analysis}
\label{sec::error_analysis}
\begin{table*}[t!]
    \centering
    \small
    \resizebox{\textwidth}{!}{
    \begin{tabular}{p{0.5\textwidth}p{0.5\textwidth}}
    \toprule
    Example 1~(PEP misleads implicit meaning) & Example 2~(PEP discards important information)\\
    \midrule
    \textbf{Q}: An interior design firm offers installation for \$129.00.  It includes hanging 4 mirrors, 2 shelves, 1 chandelier, and 10 pictures.  They will install additional items for an extra \$15.00 per item.  Angela has 6 mirrors and 2 chandeliers and 20 pictures \textcolor{cyan}{that she needs hung}.  How much will this cost her? & 
    \textbf{Q}: Mrs. Harrington bought 12 boxes of oranges at the market. She gave her mom and her sister 2 boxes of oranges each. \textcolor{cyan}{Then she kept 1/4 of the oranges} and sold the rest. How many oranges did Mrs. Harrington sell if each box contains 20 oranges?\\
    \midrule
    \textbf{PEP + CoT}: & \textbf{PEP + CoT}:  \\
    S1: What is the base installation fee for the interior design firm?
    \$129.00.
    S2: What is included in the base installation fee?
    4 mirrors, 2 shelves, 1 chandelier, and 10 pictures.
    S3: How much does the interior design firm charge for each additional item installed?
    Extra \$15.00 per item installed.
    S4: How many additional items does Angela need installed? Angela needs 6 mirrors, 2 chandeliers, and 20 pictures installed, \textcolor{red}{which is a total of 28 additional items}.
    
    Solution:
    The cost of the additional items is 28 items x \$15.00 per item = \$420.00. Therefore, the total cost for Angela's installation is \$129.00 + \$420.00 = \$549.00. 

    &
    S1: How many boxes of oranges did Mrs. Harrington buy at the market? 12 boxes. S2: How many boxes of oranges did Mrs. Harrington give to her mom and her sister? 2 boxes. S3: How many boxes of oranges did Mrs. Harrington keep for herself?
    \textcolor{red}{Mrs. Harrington kept 1/4 of the oranges.}S4: How many boxes of oranges did Mrs. Harrington sell? The rest. 
    
    Solution: 1. Calculate the total number of oranges Mrs. Harrington bought: 12x20=240. 2. Calculate the total number of oranges she gave to her mom and her sister: 2x2=4, 4x20=80. 3. Total number of oranges kept for herself: \textcolor{red}{1/4x240=60}. 4. Calculate the total number of oranges sold: 240-80-60=100.
    \\
    \bottomrule
    \end{tabular}}
    \caption{Error Analysis via \texttt{turbo}. When pre-processing the problems, PEP may changes the underline meaning and discarding information~(\textcolor{red}{red}) of the original sentence~(\textcolor{cyan}{cyan}).}
    \label{tab::errors}
\end{table*}
We present two error cases in Table~\ref{tab::errors}. PEP may ignore the potential implicatures of original sentence, resulting in ambiguity of rephrasing. It may also focus too much on the local clause and neglect the nested logical structure and temporal relations for given statements.

Besides, PEP might break the continuous context, thus changing the implicit meanings. The focus on localities might also constrain required associative thinking. We present these error cases in Sec~\ref{sec::C} in Appendix. In addition, except increasing the cost of context length, PEP may also be inefficient for very long descriptions. For certain forms of data, such as short but challenging questions, structured data in table, it could be difficult to elaborate. 

\section{Conclusion}
In this study, we proposed a novel method, Problem Elaboration Prompting (PEP), to improve the inference capabilities of LLMs. PEP offers several advantages: 1) PEP outperforms baselines across mathematical datasets, decoding strategies, and answer types; 2) PEP does not necessitate the complex creation of plans or sub-questions, but just echoes and enriches the problem context in one pass. It is also compatible with most prompting methods that enhance prefix-prompts or rationales; 3) PEP helps mitigate the distraction issue, indicating its potential in tackling other types of ill-formed problems.

\bibliographystyle{colm2024_conference}
\bibliography{custom}

\appendix
\section{Experiment Details}
\label{sec::prompt-selection}

\begin{table}[h]
\centering
\resizebox{\linewidth}{!}{
\begin{tabular}{cp{.8\linewidth}c}
\toprule
Label & \centering Prompt & Accuracy \\ \midrule
C0 & Think it step by step. & 80.0 \\
C1 & Solve the question step by step & 84.0 \\ \midrule
P1 & Decompose the given question into smaller segments, elucidating each segment as you rephrase it. & 87.0 \\
P2 & Break down the following question into concise phrases and elaborate on each phrase while rewriting. & 86.0 \\
P3 & Rewrite the following question by decomposing it into shorter clauses and providing explanations for each clause. & 85.0 \\
P4 & Restructure the subsequent question by dissecting it into more concise clauses and enhancing clarity through explanatory rephrasing. & 80.0 \\ \midrule
P5 & Break down the problem into independent, concise, and complete phrases, aligning the meaning of each phrase with the original text. Focus on expressing only one concept, action, or condition in each phrase. Then, provide detailed explanations for each phrase, analyzing the implicit messages, defining terms, and using precise professional vocabulary to accurately convey the meaning, aiming to match the potential intention of the target problem. & 62.0 \\ \bottomrule
\end{tabular}
}
\caption{Results of prompt selection. While P1 was selected and employed in our experiments to demonstrate the effectiveness of the concept of elaboration, there could be better prompts.}
\label{tab::prompt-selection}
\end{table}

\paragraph{Prompt Design and Selection.} We randomly sampled 200 points from GSM8k dataset for prompt selections. The alternatives are generated by communicating with ChatGPT. We test the original zero-shot instruction and the standard CoT used in our experiments. Then, we test four brief prompts and one particularly detailed prompt. The results are list in Table~\ref{tab::prompt-selection}. An interesting phenomenon is that the overly detailed prompt~(P5) resulted in a significant decrease.

\paragraph{Model Usage.} We used \texttt{turbo-0301}, an early version of ChatGPT, for instruction selection and type analysis in PEP. Given the possibility that OpenAI may restrict access to early models, we tested PEP on four open-source models, \texttt{text-davinci-003} and the recently released \texttt{turbo-0125}, as well to validate the generalization of the instructions. We utilized both \texttt{Llamma} models and \texttt{Mistral-7B} with bfloat16. Due to cost constraints, we loaded \texttt{Mistral-8x7B} in 4 bits.

\section{All used instructions and exemplars}
\label{sec::prompt}

\subsection{Zero-shot prompts}

To ensure fairness and clarity of semantics when combining multiple instructions, we use ``Let's solve the question step by step'' as the zero-shot CoT instead of ``Let's think step by step.''. We compared these two instructions in Table~\ref{tab::prompt-selection}. The used prompts are list as follow:

\prompt{CoT = "Let's solve the problem step by step. \{IRR\_Inst\}\{FORMAT\_Inst\}\textbackslash nQuestion: \{qst\}" \\
}

\prompt{PEP = "Decompose the given question into smaller segments, elucidating each segment as you rephrase it. Then, solve the problem step by step. \{IRR\_Inst\}\{FORMAT\_Inst\}\textbackslash nQuestion: \{qst\}" \\
}

\prompt{IRR\_Inst = "Feel free to ignore irrelevant information given in the questions." \\
}

\subsection{Extract the answers}

In order to extract results better, we add a standardized output instruction after each prompt during generation, namely \\

 (1) for free-answered questions, we use 

\prompt{FORMAT\_Inst = "End the solution in the format: 'Final answer: \textbackslash boxed\{X\}', where X is arabic numerals or 'N\textbackslash A' if the problem is unsolvable."\\}

(2) for questions with options, we use 

\prompt{FORMAT\_Inst = "End the solution in the format: 'Final answer: \textbackslash boxed\{X\}', where X is the choice."\\}

Finally, we use a one-shot exemplars to extract the answers from generations by \texttt{turbo-0125}. For those unrecognized solutions, we extract the answers manually.

\prompt{Extract\_Template = ```Given the textual solution or code execution solution, output the numeric answer that can be converted into float value of the problem. If the solution does not yield a result, output "unsolved". Only output the numeric value or "unsolved". \\\\
\#\#\# Example: \\
Solution: The total amount of money Janet makes from selling eggs at the farmers' market per day is 21 - 4 = 17 eggs x \$2 = \$34. \\ Therefore, the final answer is: Janet makes \$34 per day at the farmers' market. \\Answer: \$34 \\
Result: 34 \\
\#\#\# \\
Original Problem: \{qst\} \\
Solution: \{sol\} \\
Result:```\\}

\subsection{Few-shot exemplars}

In this section, we enumerate all the exemplars used. They are directly adopted for CoT and L2M from \cite{Zhou2022LeasttoMostPE}, which are specifically designed for GSM8k. For PEP, PaS, and Self-ask, we re-generate the exemplars using GPT-4 for proper modifications.

In practice, the one-shot setting uniformly employs the first exemplar, while the four-shot utilizes all examples. When integrating PEP with other problem-related methods, we simply append the corresponding elaboration part of PEP to the original question of other examples, before any other generations. When employing \texttt{IRR-Inst}, we position it at the beginning.

The overall template structure is as follows, using the one-shot L2M+PEP+IRR\_inst as an example for illustration:

\prompt{```Solve grade school math problems. Feel free to ignore irrelevant information given in the questions.\textbackslash n
Question: Elsa has 5 apples. Anna has 2 more apples than Elsa. How many apples do they have together?\\
Problem Elaboration:\\
Segment 1: Elsa has 5 apples.\\
Segment 2: Anna has 2 more apples than Elsa.\\
Segment 3: How many apples do they have together?\\
Rephrased question: If Elsa has 5 apples and Anna has 2 more apples than Elsa, how many apples do they have together?\\
\\
Answer: Let’s break down this rephrased problem: 1. How many apples does Anna have? 2. How many apples do Elsa and Anna have together?\\
1. Anna has 2 more apples than Elsa. So Anna has 2 + 5 = 7 apples.\\
2. Elsa and Anna have 5 + 7 = 12 apples together.\\
\\
Question: \{qst\}\\
Problem Elaboration:```\\}

\subsubsection{exemplars for CoT}

\prompt{Question: Elsa has 5 apples. Anna has 2 more apples than Elsa. How many apples do they have together? \\
Answer: Anna has 2 more apples than Elsa, so Anna has 2 + 5 = 7 apples. Elsa and Anna have 5 + 7 = 12 apples together. The answer is 12.\\ \\
Question: Four years ago, Kody was only half as old as Mohamed. If Mohamed is currently twice 30 years old, how old is Kody?\\
Answer: We were told that Mohamed is currently twice 30 years old, so he is currently 30 * 2 = 60 years old. That means that four years ago he must have been 60 - 4 = 56 years old. Four years ago, Kody was half as old as Mohamed, so Kody must have been 56 / 2 = 28 years old then. Since Kody was 28 years old four years ago, she must now be 28 + 4 = 32 years old. The answer is 32.
\\}

\prompt{Question: Carla bought 2 bags of mini peanut butter cups on clearance. Each bag was \$6.00 but was 75\% off. How much did she spend on 2 bags of candy? \\
Answer: Each bag was \$6.00 but was 75\% off. So each bag cost \$6.00 * (1 - 0.75) = \$6.00 * 0.25 = \$1.50. Carla bought 2 bags. So she spent \$1.50 * 2 = \$3.00. The answer is 3.\\}

\prompt{Question: If Pam is currently twice as young as Rena is, and in 10 years Rena will be 5 years older than her, how old is Pam now?\\
Answer: Since Rena will be 5 years older than Pam in 10 years, she must be 5 years older than Pam now as well. If Pam is currently twice as young as Rena, that means that Rena is currently twice as old as Pam is. So if P stands for Pam’s age now and R stands for Rena’s age now, then we know that R = 2 * P And since Rena is 5 years older than Pam now, we know that R = P + 5. By substitution, we have P + 5 = 2 * P, which means that P = 5. The answer is 5.\\}

\prompt{FOUR\_SHOT\_CoT = ```\{exemplars\}\\
Question: \{qst\}\\
Answer: \\}

\subsubsection{exemplars for PEP}

\prompt{Question: Elsa has 5 apples. Anna has 2 more apples than Elsa. How many apples do they have together?\\
Problem Elaboration: \\
Segment 1: Elsa has 5 apples. This segment tells us the number of apples Elsa has. \\
Segment 2: Anna has 2 more apples than Elsa. This segment tells us that Anna has more apples than Elsa, specifically 2 more. \\
Segment 3: How many apples do they have together? This segment is asking us to find the total number of apples both Elsa and Anna have combined.\\
Solution: \\
Step 1: Determine the number of apples Elsa has. Elsa has 5 apples. \\
Step 2: Determine the number of apples Anna has. Anna has 2 more apples than Elsa, so she has 5 + 2 = 7 apples.\\
Step 3: Determine the total number of apples they have together. Together, Elsa and Anna have 5 + 7 = 12 apples. The answer is 12.\\}

\prompt{Question: Four years ago, Kody was only half as old as Mohamed. If Mohamed is currently twice 30 years old, how old is Kody?\\
Problem Elaboration: \\
Segment 1: Four years ago, Kody was only half as old as Mohamed. This means that the age difference between Kody and Mohamed is constant and it is the same four years ago as it is now. \\
Segment 2: If Mohamed is currently twice 30 years old. This means that Mohamed's current age is 60 years old.\\
Solution:\\
Step 1: Determine Mohamed's age four years ago. If Mohamed is currently 60 years old, then four years ago he was 60 - 4 = 56 years old.\\
Step 2: Determine Kody's age four years ago. Since Kody was half as old as Mohamed four years ago, then Kody was 56 / 2 = 28 years old four years ago.\\
Step 3: Determine Kody's current age. If Kody was 28 years old four years ago, then Kody is currently 28 + 4 = 32 years old. The answer is 32.\\}

\prompt{Question:  Carla bought 2 bags of mini peanut butter cups on clearance. Each bag was \$6.00 but was 75\% off. How much did she spend on 2 bags of candy?\\
Problem Elaboration:\\ 
Segment 1: Identify the original price of the bags of candy. The original price of each bag of candy is \$6.00.\\
Segment 2: Determine the discount on each bag. The bags are 75\% off. \\
Segment 3: Calculate the discounted price of each bag. To find the discounted price, we need to calculate 75\% of \$6.00. \\
Segment 4: Determine the total cost for 2 bags. Once we have the discounted price of one bag, we multiply it by 2 to find the total cost for 2 bags.\\
Solution:
Step 1: The original price of each bag is \$6.00.
Step 2: The discount on each bag is 75\%. 
Step 3: To calculate 75\% of \$6.00, we multiply 6 by 0.75, which equals \$4.50. This means that \$4.50 is the amount of the discount.
Step 4: To find the discounted price of each bag, we subtract the discount from the original price. So, \$6.00 - \$4.50 = \$1.50. Each bag costs \$1.50 after the discount.
Step 5: To find the total cost for 2 bags, we multiply the discounted price by 2. So, \$1.50 * 2 = \$3.00. The answer is 3.\\}

\prompt{Question: If Pam is currently twice as young as Rena is, and in 10 years Rena will be 5 years older than her, how old is Pam now?\\
Problem Elaboration: \\
Segment 1: Pam is currently twice as young as Rena is. This means that Pam's current age is half of Rena's current age.\\
Segment 2: In 10 years, Rena will be 5 years older than Pam. This means that if we add 10 years to both Pam's and Rena's current ages, the difference between their ages will be 5 years.\\
Solution:\\
Step 1: Let's denote Rena's current age as R and Pam's current age as P. From the first segment, we know that P = R/2.\\
Step 2: From the second segment, we know that R + 10 = P + 10 + 5. We can simplify this to R = P + 5.\\
Step 3: Now we can substitute P from the first equation into the second equation. So, R = R/2 + 5.\\
Step 4: To solve for R, we multiply both sides of the equation by 2 to get rid of the fraction. This gives us 2R = R + 10.\\
Step 5: Subtract R from both sides to get R = 10. So, Rena is currently 10 years old.\\
Step 6: Substitute R = 10 into the first equation to find P. This gives us P = 10/2 = 5. So, Pam is currently 5 years old. The answer is 5.\\}

\prompt{FOUR\_SHOT\_PEP = ```\{exemplars\}\\
\\
Question: \{qst\}\\
Problem Elaboration: ```}

\subsubsection{exemplars for L2M}

\prompt{Question: Elsa has 5 apples. Anna has 2 more apples than Elsa. How many apples do they have together?\\
Answer: Let’s break down this problem: 1. How many apples does Anna have? 2. How many apples do Elsa and Anna have together?\\
1. Anna has 2 more apples than Elsa. So Anna has 2 + 5 = 7 apples.\\
2. Elsa and Anna have 5 + 7 = 12 apples together. The answer is 12.\\}

\prompt{Question: Four years ago, Kody was only half as old as Mohamed. If Mohamed is currently twice 30 years old, how old is Kody?\\
Answer: Let’s break down this problem: 1. How old was Mohamed four years ago? 2. How old is Kody?\\
1. We were told that Mohamed is currently twice 30 years old, so he is currently 30 * 2 = 60 years old. That means that four years ago he must have been 60 - 4 = 56 years old.\\
2. Four years ago, Kody was half as old as Mohamed, so Kody must have been 56 / 2 = 28 years old then. Since Kody was 28 years old four years ago, she must now be 28 + 4 = 32 years old. The answer is 32.\\}

\prompt{Question:  Carla bought 2 bags of mini peanut butter cups on clearance. Each bag was \$6.00 but was 75\% off. How much did she spend on 2 bags of candy?\\
Answer: Let’s break down this problem: 1. How much did she spend on 2 bags of candy?\\
1. Each bag was \$6.00 but was 75\% off. So each bag cost \$6.00 * (1 - 0.75) = \$6.00 * 0.25 = \$1.50. Carla bought 2 bags. So she spent \$1.50 * 2 = \$3.00. The answer is 3.\\}

\prompt{Question: If Pam is currently twice as young as Rena is, and in 10 years Rena will be 5 years older than her, how old is Pam now?\\
Answer: Let’s break down this problem: 1. How much older is Rena than Pam currently? 2. How old is Pam now?\\
1. Since Rena will be 5 years older than Pam in 10 years, she must be 5 years older than Pam now as well.\\
2. If Pam is currently twice as young as Rena, that means that Rena is currently twice as old as Pam is. So if P stands for Pam’s age now and R stands for Rena’s age now, then we know that R =2 * P And since Rena is 5 years older than Pam now, we know that R = P + 5. By substitution, we
have P + 5 = 2 * P, which means that P = 5. The answer is 5.\\}

\prompt{FOUR\_SHOT\_L2M = ```\{exemplars\}\\
Question: \{qst\}\\
Answer: Let’s break down this problem:```\\}

\subsubsection{exemplars for Self-ask}

\prompt{Question: Elsa has 5 apples. Anna has 2 more apples than Elsa. How many apples do they have together?\\
Are follow up questions needed here: Yes.\\
Follow up: How many apples does Anna have?\\
Intermediate answer: Anna has 5 + 2 = 7 apples.\\
Follow up: How many apples do Elsa and Anna have together?\\
Intermediate answer: Elsa and Anna have 5 + 7 = 12 apples together. The answer is 12.\\}

\prompt{Question: Four years ago, Kody was only half as old as Mohamed. If Mohamed is currently twice 30 years old, how old is Kody? \\
Are follow up questions needed here: Yes.\\
Follow up: How old is Mohamed currently?\\
Intermediate answer: Mohamed is 30 * 2 = 60 years old.
Follow up: How old was Kody four years ago?\\
Intermediate answer: Kody was (60 - 4) / 2 = 28 years old four years ago.\\
So the final answer is: Kody is 28 + 4 = 32 years old. The answer is 32.\\}

\prompt{Question: Carla bought 2 bags of mini peanut butter cups on clearance. Each bag was \$6.00 but was 75\% off. How much did she spend on 2 bags of candy?\\
Are follow up questions needed here: Yes.\\
Follow up: How much was the discount for each bag?\\
Intermediate answer: The discount for each bag is \$6.00 * 75\% = \$4.50.\\
Follow up: How much did Carla pay for each bag after the discount?\\
Intermediate answer: Carla paid \$6.00 - \$4.50 = \$1.50 for each bag.\\
So the final answer is: Carla spent \$1.50 * 2 = \$3.00 on 2 bags of candy. The answer is 3.00.\\}

\prompt{Question: If Pam is currently twice as young as Rena is, and in 10 years Rena will be 5 years older than her, how old is Pam now?\\
Are follow up questions needed here: Yes.\\
Follow up: What about Rena and Pam's current ages?\\
Intermediate answer: It tells us that Rena's age is twice Pam's age. So if P stands for Pam’s age now and R for Rena’s age now, then R = 2 * P. And since Rena is 5 years older than Pam now, we have R = P + 5.\\
Follow up: What is Pam’s age now?\\
Final answer: By substituting P + 5 in place of R in equation R = 2 * P, we get P + 5 = 2 * P, which simplifies to P = 5. So, Pam is 5 years old. The answer is 5.\\}

\prompt{FOUR\_SHOT\_SK =```\{exemplars\}\\
\\
Question: \{qst\}\\
Are follow up questions needed here: Yes.\\
Follow up: ```\\}

It's worth noting that the template recommended in self-ask~\citep{Press2022MeasuringAN} actually ends with ``Are follow up questions needed here:'', but we found it always generates a ``Yes'' or ``No'' and then stops, especially in one-shot setting. Therefore, we made above adjustments.

\subsubsection{exemplars of PaS}

\prompt{Question: Elsa has 5 apples. Anna has 2 more apples than Elsa. How many apples do they have together?\\
Plan:\\
Step 1: Find out how many apples Anna has.  \\
Step 2: Add the number of apples Elsa has to the number of apples Anna has to find the total number of apples they have together.\\
Solution:\\
Step 1: Anna has 5 + 2 = 7 apples.\\
Step 2: Together, Elsa and Anna have 5 (Elsa's apples) + 7 (Anna's apples) = 12 apples. So, Elsa and Anna have 12 apples together. The answer is 12.\\}

\prompt{Question: Four years ago, Kody was only half as old as Mohamed. If Mohamed is currently twice 30 years old, how old is Kody? \\
Plan: \\
Step 1: Find out how old Mohamed is currently.\\
Step 2: Find out how old Mohamed was four years ago.\\
Step 3: Since Kody was half as old as Mohamed four years ago, find out Kody's age four years ago.\\
Step 4: Add four years to Kody's age to find out his current age. \\
Solution:\\
Step 1: Mohamed is currently 2 * 30 = 60 years old.\\
Step 2: Four years ago, Mohamed was 60 - 4 = 56 years old.\\
Step 3: Four years ago, Kody was 56 / 2 = 28 years old.
Step 4: Currently, Kody is 28 + 4 = 32 years old. So, Kody is 32 years old. The answer is 32.\\}

\prompt{Question: Carla bought 2 bags of mini peanut butter cups on clearance. Each bag was \$6.00 but was 75\% off. How much did she spend on 2 bags of candy?\\
Plan:\\
Step 1: Find out how much discount Carla got on each bag.\\
Step 2: Subtract the discount from the original price to find out the price Carla paid for each bag.
Step 3: Multiply the price Carla paid for each bag by the number of bags she bought to find out how much she spent in total.\\
Solution:\\
Step 1: The discount on each bag is 75/100 * \$6.00 = \$4.50.\\
Step 2: The price Carla paid for each bag is \$6.00 - \$4.50 = \$1.50.\\
Step 3: Carla spent \$1.50 * 2 = \$3.00 on 2 bags of candy. So, Carla spent \$3.00 on 2 bags of candy. The answer is 3.00.\\}

\prompt{Question: If Pam is currently twice as young as Rena is, and in 10 years Rena will be 5 years older than her, how old is Pam now?\\
Plan: \\
Step 1: Set up an equation based on the information that Rena's age is twice Pam's age.\\
Step 2: Set up another equation based on the information that Rena is 5 years older than Pam.\\
Step 3: Substitute the second equation into the first to solve for Pam's age.\\
Solution:\\
Step 1: If P stands for Pam’s age now and R for Rena’s age now, then R = 2 * P.\\
Step 2: And since Rena is 5 years older than Pam now, we have R = P + 5.\\
Step 3: By substituting P + 5 in place of R in equation R = 2 * P, we get P + 5 = 2 * P, which simplifies to P = 5. So, Pam is 5 years old. The answer is 5.\\}

\prompt{FOUR\_SHOT\_PaS = ```\{exemplars\}\\
\\
Question: \{qst\}\\
Plan:```\\}

\subsection{Implementation of PoT}
We implement PoT using the following template. For PEP, we insert null into \{ela\}. For the PoT+PEP, we first use zero-shot PEP to generate the content for the Elaboration part, and then insert it into \{ela\}.

\prompt{PoT\_zeroshot\_temp = ```\# Question: \{qst\} \\
\{ela\} \\
\\
\# Answer the question by implementing a solution() function.  \\
\# Generate the code only.
\\
\# Let's write a Python program step by step, and then return the answer \\
\# Firstly, we need write the solution() starting with defining variable:```\\}

\section{Example Input and Output Pairs}
\label{sec::C}

\begin{table*}[t!]
    \centering
    \small
    \resizebox{\textwidth}{!}{
    \begin{tabular}{cp{0.93\textwidth}}
    \toprule
    \textbf{Question 1}: & 
    It is approximately 1955 kilometers from San Diego, California to New York City, New York. If \textcolor{cyan}{Bernice drove 325 kilometers for 4 days}, how many kilometers will she still need to drive? 
    \\ \midrule
    \textbf{CoT} &
    To find the total distance Bernice drove, we can multiply the distance she drove each day (325 kilometers) by the number of days she drove (4 days): \textcolor{blue}{325 km/day x 4 days = 1300 kilometers}. Therefore, Bernice drove 1300 kilometers. To find how many kilometers she still needs to drive, we can subtract the distance she drove from the total distance between San Diego and New York City: 1955 km - 1300 km = 655 kilometers. Therefore, Bernice still needs to drive 655 kilometers. (\textbf{Correct})
    \\ \midrule
    \multicolumn{2}{p{\textwidth}}{ \textbf{Error: The decomposition broke the continuous context into segments, changing the related implicit implicatures.}}\\
    
    \textbf{PEP + CoT}:    
    &
    Segment 1: What is the distance between San Diego, California and New York City, New York? The distance between San Diego, California and New York City, New York is approximately 1955 kilometers. Segment 2: \textcolor{red}{How many kilometers did Bernice drive? Bernice drove 325 kilometers}. Segment 3: How many days did Bernice drive for? Bernice drove for 4 days. Segment 4: How many kilometers will Bernice still need to drive? To find out how many kilometers Bernice will still need to drive, we need to subtract the distance she has already driven from the total distance between San Diego and New York. \textcolor{red}{1955 km - 325 km = 1630 km}. Therefore, Bernice will still need to drive 1630 kilometers.
    \\ \midrule
    \textbf{Question 2}: & 
    \textcolor{cyan}{Sandy wants to lose as much weight as Joey does} but needs 4 weeks to lose the same amount of weight that Joey loses in a single week. If Joey loses 8 pounds in 4 weeks, how many weeks will it take Sandy to lose the same amount of weight?
    \\ \midrule
    \textbf{CoT} & 
    First, we need to find out how much weight Joey loses in a single week. We can do this by dividing the total weight he loses in 4 weeks by 4: 8 pounds ÷ 4 weeks = 2 pounds per week.
    So Joey loses 2 pounds per week. Next, we need to figure out how much weight Sandy wants to lose. \textcolor{blue}{We know she wants to lose the same amount as Joey, which is 8 pounds.} Now we can use the information given in the problem to find out how many weeks it will take Sandy to lose 8 pounds. We know that Sandy needs 4 weeks to lose the same amount of weight that Joey loses in a single week. So we can set up a proportion: 4 weeks to lose 2 pounds = x weeks to lose 8 pounds. To solve for x, we can cross-multiply: 4 weeks * 8 pounds = 2 pounds * x weeks; 32 pounds = 2x; Divide both sides by 2: 16 weeks = x. So it will take Sandy 16 weeks to lose the same amount of weight that Joey loses in 4 weeks.~(\textbf{Correct})
    \\ \midrule
    \multicolumn{2}{p{\textwidth}}{ \textbf{Error: The elucidation emphasized and rephrased the given condition, changing the underlying implicatures.}}\\
    \textbf{PEP + CoT}: &
    Segment 1: Sandy wants to lose as much weight as Joey does.
    \textcolor{red}{Elucidation: Sandy has a weight loss goal that is equivalent to Joey's weight loss goal.}
    Segment 2: Sandy needs 4 weeks to lose the same amount of weight that Joey loses in a single week.
    Elucidation: Sandy's weight loss rate is slower than Joey's weight loss rate. It takes Sandy 4 weeks to lose the same amount of weight that Joey loses in just one week.
    Segment 3: Joey loses 8 pounds in 4 weeks.
    Elucidation: Joey's weight loss goal is to lose 8 pounds, and he achieves this goal in 4 weeks.
    Segment 4: How many weeks will it take Sandy to lose the same amount of weight?
    Elucidation: The question is asking for the number of weeks it will take Sandy to achieve the same weight loss goal as Joey.
    
    Solution: Since Joey loses 8 pounds in 4 weeks, his weight loss rate is 8/4 = 2 pounds per week. Since Sandy needs 4 weeks to lose the same amount of weight that Joey loses in a single week, her weight loss rate is 1/4 = 0.25 pounds per week. \textcolor{red}{To lose 8 pounds like Joey, Sandy will need to lose at the same rate of 2 pounds per week.} Therefore, it will take Sandy 8/2 = 4 weeks to lose the same amount of weight as Joey.
    \\\midrule
    \textbf{Question 3}: & 
    Patty's Plumbing charges \$40 to visit a house to make a repair, plus \$35 per hour, or part thereof, for labor, plus parts. \textcolor{cyan}{One job took 2.25 hours} and used \$60 in parts. How much did Patty charge?
    \\ \midrule
    \textbf{CoT} & 
    1. Calculate the labor cost: The job took 2.25 hours, \textcolor{blue}{so Patty charged for 3 hours} (since she charges for any part of an hour). The labor cost is \$35 per hour, so the labor cost for 3 hours is 3 x \$35 = \$105. 2. Calculate the total cost: The visit fee is \$40. The labor cost is \$105. The parts cost is \$60. The total cost is \$40 + \$105 + \$60 = \$205. Therefore, Patty charged \$205 for the job.~(\textbf{Correct})
    \\ \midrule
    \multicolumn{2}{p{\textwidth}}{ \textbf{Error: PEP focuses on known information and may constrain the associative thinking of CoT during reasoning.}}\\
    \textbf{PEP + CoT}: &
    Segment 1: What is the base fee for Patty's Plumbing to visit a house for a repair? Patty's Plumbing charges \$40 to visit a house for a repair. Segment 2: How much does Patty's Plumbing charge for labor? Patty's Plumbing charges \$35 per hour, or part thereof, for labor. Segment 3: How much does Patty's Plumbing charge for parts? The job used \$60 in parts. Segment 4: How long did the job take? The job took 2.25 hours.
    
    Solution: To calculate the total cost of the job, we need to add the base fee, labor cost, and parts cost. Base fee: \$40
    Labor cost: \textcolor{red}{\$35 per hour x 2.25 hours = \$78.75} Parts cost: \$60. Total cost: \$40 + \$78.75 + \$60 = \$178.75. Therefore, Patty charged \$178.75 for the job.
    \\
    \bottomrule
    \end{tabular}}
    \caption{Error cases from GSM8k. For certain sentences in the problem~(\textcolor{cyan}{cyan}), CoT can understand it correctly~(\textcolor{blue}{blue}), while PEP may lead to erroneous reasoning and generate errors~(\textcolor{red}{red}).}
    \label{tab::more-errors}
\end{table*}

In Table~\ref{tab::more-errors} we provide more error cases. As shown, LLMs could also make mistakes during elaboration, misleading the following reasoning. PEP may break down the continuous context, thus changing the implicit implicatures. Besides, PEP may focus too much on the locality of sentences, constraining the associative thinking of CoT.

\end{document}